\documentstyle[11pt,colacl]{article}

\title{Language Identification With Confidence Limits}
\author{David Elworthy\\Canon Research Centre Europe\\1 Occam Court\\Occam
Road\\Surrey Research Park\\Guildford GU2 5YJ\\United Kingdom\\{\tt dahe@cre.canon.co.uk}}

\begin{document}
\maketitle

\begin{abstract}
A statistical classification algorithm and its application to language
identification from noisy input are described. The main innovation is to
compute confidence limits on the classification, so that the algorithm
terminates when enough evidence to make a clear decision has been made, and so
avoiding problems with categories that have similar characteristics. A second
application, to genre identification, is briefly examined. The results show
that some of the problems of other language identification techniques can be
avoided, and illustrate a more important point: that a statistical language
process can be used to provide feedback about its own success rate.
\end{abstract}

\bibliographystyle{acl}
\section{Introduction}

Language identification is an example of a general class of problems in which
we want to assign an input data stream to one of several categories as quickly
and accurately as possible. It can be solved using many techniques, including
knowledge-poor statistical approaches. Typically, the distribution of n-grams
of characters or other objects is used to form a model. A comparison of the
input against the model determines the language which matches best. Versions
of this simple technique can be found in Dunning~(1994) and Cavnar and
Trenkle~(1994), while an interesting practical implementation is described by
Adams and Resnik~(1997).

A variant of the problem is considered by Sibun and Spitz~(1994), and Sibun
and Reynar~(1996), who look at it from the point of view of Optical Character
Recognition (OCR). Here, the language model for the OCR system cannot be
selected until the language has been identified. They therefore work with
so-called shape tokens, which give a very approximate encoding of the
characters' shapes on the printed page without needing full-scale OCR. For
example, all upper case letters are treated as being one character shape, all
characters with a descender are another, and so on. Sequences of character
shape codes separated by white space are assembled into word shape
tokens. Sibun and Spitz then determine the language on the basis of linear
discriminant analysis (LDA) over word shape tokens, while Sibun and Reynar
explore the use of entropy relative to training data for character shape
unigrams, bigrams and trigrams. Both techniques are capable of over 90\%
accuracy for most languages. However, the LDA-based technique tends to perform
significantly worse for languages which are similar to one another, such as
the Norse languages. Relative entropy performs better, but still has some
noticeable error clusters, such as confusion between Croatian, Serbian and
Slovenian.

What these techniques lack is a measure of when enough information has been
accumulated to distinguish one language from another reliably: they examine
all of the input data and then make the decision. Here we will look at a
different approach which attempts to overcome this by maintaining a measure of
the total evidence accumulated for each language and how much confidence there
is in the measure. To outline the approach:
\begin{enumerate}
\item The input is processed one (word shape) token at a time. For each
language, we determine the probability that the token is in that language,
expressed as a 95\% confidence range.
\item The values for each word are accumulated into an overall score with a
confidence range for the input to date, and compared both to an absolute
threshold, and with each other. Thus, to select a language, we require not
only that it has a high score (probability, roughly), but also that it is
significantly better scoring than any other.
\item If the process fails to make a decision on the data that is available,
the subset of the languages which have exceeded the absolute threshold can be
output, so that even if a final decision has not been made, the likely
possibilities have been narrowed down.
\end{enumerate}
We look at this procedure in more detail below, with particular emphasis on
how the underlying statistical model provides confidence
intervals. An evaluation of the technique on data similar to that used by
Sibun and Reynar follows\footnote{Some ideas related to the use of confidence
limits can also be found in Dagan {\it et al.}~(1991), applied in a different
area.}.

\section{The Identification Algorithm}

The essential idea behind the identification algorithm is to accumulate the
probability of the language given the input tokens for each language, treating
each token as an independent event. To obtain the probability of a language
$l$ given a token $t$, $p(l|t)$, we use Bayes' rule:
\[ p(l|t) = \frac{p(t|l) p(l)}{p(t)} \]
where $p(t|l)$ is the probability of the token if the language is known,
$p(t)$ is the {\em a priori} probability of the token, and $p(l)$ is the {\em
a priori} probability of the language. We will assume that $p(l)$ is constant
(all languages are equi-probable) and drop it from the computation; in the 
tests, we will use the same amount of training data for each language. The
other two terms are estimated from training data, using the procedure
described in section~\ref{train}.

\subsection{The language model and the algorithm}

The input to the algorithm consists of a stream of tokens, such as word shape
tokens (as in Sibun and Spitz, or Sibun and Reynar) or words themselves. The
model for each language contains the probability of each known token given the
language, expressed as three values: the basic probability, and the lower and
upper limits of a range containing this probability for a specific level of
confidence. We will denote these by $p_{B}(t|l)$, $p_{L}(t|l)$, $p_{H}(t|l)$,
for base, low and high values. The probability that a token which has never
been seen before is in a language is also present in the model of the
language. In addition, there is a language independent model, containing the
$p(t)$ values. No confidence range is used for them, although this would be a
simple extension of the technique.

The algorithm proceeds by processing tokens, building up evidence about each
language in three accumulators. The accumulators represent the overall
probability of the language given the entire stream of tokens to date, again
as base, low and high values, denoted $a_{B}(l)$, $a_{L}(l)$, $a_{H}(l)$. They
are set to zero at the start of processing, and the logarithms of the
probabilities are added to them as each token is processed. By taking
logarithms of probabilities, we are in effect measuring the amount of evidence
for each language, expressed as information content. From a practical point of
view, using logarithms also helps keep all the values in a reasonable range
and so avoids numerical underflow.

After processing each token, two tests are applied. Firstly, we examine the
base accumulator for the language which has the highest accumulated total, and
test whether it is greater than a fixed threshold, called the activation
threshold. If it is, then we conclude that enough information has been
accumulated to try to make a decision. The low value for this language
$a_{L}(l)$ is then compared against the high value $a_{H}(l')$ for the next
best language $l'$, and if $a_{L}(l)$ exceeds $a_{H}(l')$ language $l$ is
output and the algorithm halts. Otherwise, the process continues with the next
token, until the best choice language is a clear ``winner'' over any other.

Finally, if we reach the end of the input data without a decision being made,
several options are possible, depending on the needs of the application. We can
simply output the language with the highest base score, even if the second
test is not satisfied. Alternatively, we can output the highest scoring
language, and all other languages whose high probability is greater than the
low probability of this language.

\subsection{Training the model}
\label{train}

The model is trained using a collection of corpora for which the correct
language is known. For a given language $l$ and token $t$, let $f(t,l)$ be the
count of the token in that language and $f(l)$ be the total count of all tokens
in that language. $f(t)$ is the count of the token $t$ across all the
languages, and $F$ the count of all tokens across all languages. The
probability of the token occurring in the language $p(t|l)$ is then calculated
by assuming that the probabilities follow a binomial distribution. The idea
here is that token occurrences are binary ``events'' which are either the
given token $t$ or are not. For large $f(t,l)$, the underlying probability
can be calculated by using the normal approximation to the binomial, giving
the base probability
\[p_{B}(t|l) = \frac{f(t,l)}{f(l)}\]
The standard deviation of this quantity is
\[\sigma(t,l) = \sqrt{f(l)p_{B}(t|l)(1-p_{B}(t|l))}\]
The low and high probabilities are found by taking a given number of standard
deviations $d$ from the base probability.
\[p_{L}(t|l) = \frac{f(t,l)-d\sigma(t,l)}{f(l)}\]
\[p_{H}(t|l) = \frac{f(t,l)+d\sigma(t,l)}{f(l)}\]
In the evaluation below, $d$ was set to 2, giving 95\% confidence limits.

For lower values of $f(t,l)$, the calculation of the low and high
probabilities can be made more exact, by substituting them for the base
probability in the calculation of the standard deviation, giving
\[p_{L}(t|l) = \frac{f(t,l)-d\sqrt{f(l)p_{L}(t|l)(1-p_{L}(t|l))}}{f(l)}\]
\[p_{H}(t|l) = \frac{f(t,l)+d\sqrt{f(l)p_{H}(t|l)(1-p_{H}(t|l))}}{f(l)}\]
Approximating $1-p_{L}(t|l)$ and $1-p_{H}(t|l)$ to 1 on the grounds that the
probabilities are small, and solving the equations gives
\[p_{L}(t|l) = \frac{(\sqrt{d^{2}+4f(t,l)}-d)^{2}}{4f(l)} \]
\[p_{H}(t|l) = \frac{(\sqrt{d^{2}+4f(t,l)}+d)^{2}}{4f(l)} \]
The calculation requires marginally more computational effort than the first
case, and in practice we use it for all but very large values of $f(t,l)$,
where the approximation of $1-p_{L}(t|l)$ and $1-p_{H}(t|l)$ to 1 would break
down.

For very small values of $f(t,l)$, say less than 10, the normal
approximation is not good enough, and we calculate the probabilities by
reference to the binomial equation for the probability of $m$ (=$f(t,l)$)
successes in $n$ (= $f(l)$) trials:
\[p(m) = \frac{p^{m}(1-p)^{n-m}n!}{m!(n-m)!} \]
$p$ is the underlying probability of the distribution, and this is what we are 
after. By choosing values for $p(m)$ and solving to find $p$ we can
obtain a given confidence range. To obtain a 95\% interval, $p(m)$ is set to
0.025, 0.5 and 0.975, yielding $p_{L}(t|l)$, $p_{B}(t|l)$, and $p_{H}(t|l)$,
respectively. In fact, this is not exactly how the probability ranges for low
frequency items should be calculated: instead the cumulative probability
density function should be calculated and the range estimated from
it\footnote{Thanks to one of the referees for pointing this out.}. For the
present purposes, the low frequency items do not make much of a contribution
to the overall success rate, and so the approximation is unimportant. However,
if similar techniques were applied to problems with sparser data, then the
procedure here would have to be revised.

Finally, we need a probability for tokens which were not seen in the training
data, called the zero probability, for which we set $m=0$ in the above
equation giving
\[p(0|l) = 1-\sqrt[n]{p(m)}\]
It is not clear what it means to have a confidence measure here, and so we use
a single value for base, low and high probabilities, obtained by setting $p(m)$
to 0.95.

Similar calculations using $f(t)$ in place of $f(t,l)$ and $F$ in place of
$f(l)$ give the a priori token probabilities $p(t)$. As already noted, base,
low and high value could have been calculated in this case, but as a minor
simplification, we use only the base probability.

\section{Evaluation}

To evaluate the technique, a test was run using similar data to
Sibun and Reynar. Corpora for eighteen languages from the European Corpus
Initiative CDROM 1 were extracted and split into
non-overlapping files, one containing 2000 tokens\footnote{Sibun and Spitz,
and Sibun and Reynar, present their results in terms of lines of input, with
1-5 lines corresponding roughly to a sentence, and 10-20 lines to a
paragraph. Estimating a line as 10 words, we are therefore working with
significantly smaller data sets.}, one containing 200 tokens, and 25 files
each of 1, 5, 10 and 20 tokens. The 2000 and 200 token files
were used as training data, and the remainder for test data. Wherever possible
the texts were taken from newspaper corpora, and failing that from novels or
literature. The identification algorithm was run on each test file and the
results placed in one of four categories:
\begin{itemize}
\item Definitive, correct decision made.
\item No decision made by the end of the input, but highest scoring language
was correct.
\item No decision, highest scoring language incorrect.
\item Definitive, incorrect decision made.
\end{itemize}
The sum of the first two figures divided by the total number of tests gives a
measure of accuracy; the sum of the first and last divided by the total gives
a measure of decisiveness, expressed as the proportion of the time a
definitive decision was made. The tests were executed using word shape tokens
on the same coding scheme as Sibun and Reynar, and using the words as they
appeared in the corpus. No adjustments were made for punctuation, case,
etc. Various activation thresholds were tried: raising the threshold increases
accuracy by requiring more information before a decision is made, but reduces
decisiveness. With shapes and 2000 tokens of training data, at a
threshold of 14 or more, all the 20 token files gave 100\% accuracy. For words
themselves, the threshold was set to 22. The results of these tests appear in
table~\ref{tab1}. The figures for the activation threshold were determined by
experimenting on the data. An interesting area for further work would be to
put this aspect of the procedure on a sounder theoretical basis, perhaps by
using the {\em a priori} probabilities of the individual languages.

\begin{table*}[htb]
\begin{center}
\begin{tabular}{|c|c|c|c|c|c|c|c|c|c|c|}\hline
Test and & \multicolumn{5}{c|}{Accuracy (\%)} &
\multicolumn{5}{c|}{Decisiveness (\%)} \\ \cline{2-6} \cline{7-11}
threshold  &  \multicolumn{5}{c|}{Tokens of test data} &
\multicolumn{5}{c|}{Tokens of test data} \\ \cline{2-6} \cline{7-11}
 & 1 & 5 & 10 & 20 & {\bf All} & 1 & 5 & 10 & 20 & {\bf All} \\ \hline
Tokens (0) & 71.6 & 72.7 & 69.6 & 72.0 & {\bf 71.4} & 88.0 & 99.3 & 100 & 99.8 & {\bf 96.8}\\
Tokens (10) & 92.9 & 98.4 & 98.4 & 98.2 & {\bf 97.0} & 66.0 & 98.9 & 99.6 & 99.8 & {\bf 91.1}\\
Tokens (14) & 94.2 & 99.6 & 99.1 & 100 & {\bf 98.2} & 49.8 & 98.9 & 99.6 & 99.8 & {\bf 87.0}\\
\hline
Words (0) & 78.4 & 80.4 & 77.1 & 78.7 & {\bf 78.7} & 97.3 & 100 & 100 & 100 & {\bf 99.3}\\
Words (10) & 95.8 & 97.6 & 97.1 & 98.0 & {\bf 97.1} & 76.9 & 99.8 & 100 & 99.8 & {\bf 94.1}\\
Words (22) & 96.9 & 99.8 & 99.8 & 100 & {\bf 99.1} & 29.3 & 98.9 & 99.8 & 99.8 & {\bf 81.9}\\
\hline
\end{tabular}
\caption{Performance with 2000 tokens of training data}
\label{tab1}
\end{center}
\end{table*}
The accuracy figures are generally similar to or better than those of Sibun
and Reynar. The corresponding figures for 200 tokens of training data appear in
table~\ref{tab1.200}, for the token identification task only.
\begin{table*}[htb]
\begin{center}
\begin{tabular}{|c|c|c|c|c|c|c|c|c|c|c|}\hline
Threshold &  \multicolumn{5}{c|}{Accuracy (\%)} &
\multicolumn{5}{c|}{Decisiveness (\%)} \\ \cline{2-6} \cline{7-11}
          &  \multicolumn{5}{c|}{Tokens of test data} &
\multicolumn{5}{c|}{Tokens of test data} \\ \cline{2-6} \cline{7-11}
          & 1 & 5 & 10 & 20 & {\bf All} & 1 & 5 & 10 & 20 & {\bf All} \\ \hline
0 & 63.3 & 72.4 & 48.9 & 47.1 & {\bf 57.9} & 72.2 & 89.6 & 96.7 & 96.2 & {\bf 88.7}\\
5 & 82.2 & 88.9 & 75.6 & 75.8 & {\bf 80.6} & 58.0 & 86.4 & 93.3 & 93.8 & {\bf 82.9}\\
10 & 86.0 & 94.0 & 88.0 & 87.6 & {\bf 88.9} & 45.6 & 85.1 & 91.1 & 92.4 & {\bf 78.6}\\
\hline
\end{tabular}
\caption{Performance with 200 tokens of training data (word shape tokens only)}
\label{tab1.200}
\end{center}
\end{table*}

One of the strengths of the algorithm is that it makes a decision as soon as
one can be made reliably. Table~\ref{tab2} shows the average number of tokens
which have to be read before a decision can be made, for the cases where the
decision was correct and incorrect, and for both cases together. Again, the
results are for word shape tokens, and for words alone.
\begin{table*}[htb]
\begin{center}
\begin{tabular}{|c|c|c|c|c|c|c|}\hline
Threshold & \multicolumn{3}{c|}{Shape tokens} & \multicolumn{3}{c|}{Words} \\
\cline{2-4} \cline{5-7}
   & Correct & Incorrect & All & Correct & Incorrect & All \\ \hline
0  & 3.22 & 1.23 & 2.65 & 1.81 & 1.07 & 1.66 \\
10 & 7.33 & 4.55 & 7.28 & 5.31 & 3.88 & 5.28 \\
14 & 9.35 & 6.50 & 9.33 &      &      &      \\
22 &      &      &      & 10.6 & 8.00 & 10.6 \\
\hline
\end{tabular}
\caption{Average number of tokens read before convergence}
\label{tab2}
\end{center}
\end{table*}
The figures show that convergence usually happens within about 10 words, with
a long tailing off to the results. The longest time to convergence was 153
shape tokens.

A manual inspection of one run (2000 lines of training data, tokens,
threshold=14) shows that errors are somtimes clustered, although quite
weakly. For example, Serbian, Croatian and Slovenian show several confusions
between them, as in Sibun and Reynar's results. There are two observations to
be made here. Firstly, there are about as many other errors between these
language and languages which are unrelated to them, such as Italian, German
and Norwegian, and so the errors may be due to poor quality data rather than a
lack of discrimination in the algorithm. For example, Croatian is incorrectly
recognised as Serbian 3 times and as Slovenian once, while the languages which
are misrecognised as Croatian are German and Norwegian (once each). Secondly,
even where there are errors, the range of possibilities has been substantially
reduced, so that a more powerful process (such as full-scale OCR followed by
identification on words rather than shape tokens, or a raising of the
threshold and adding more data) could be brought in to finish the job
off. That is, the confidence limits have provided a benefit in reducing the
search space. The confusion matrix for this case appears in an appendix.

\subsection{Broader applicability}

Although the algorithm was developed with language identification in mind, it
is interesting to explore other classification problems with it. A simple and
rather crude experiment in ``genre'' identification was carried out, using the
Brown corpus. Each section of the corpus (labelled A, B, C ... R in the
original) was taken as a genre, and files of similar distribution to the
previous experiment were extracted. Because this is a more unconstrained
problem, the training set and tests sets were about 10 times the size of the
language identification task. A 20000 word file was used as training data, and
the remaining files as test data. Accuracy and decisiveness results appear in
table~\ref{tab3}.
\begin{table*}[htb]
\begin{center}
\begin{tabular}{|c|c|c|c|c|c|c|c|c|c|c|}\hline
Threshold &  \multicolumn{5}{c|}{Accuracy (\%)} &
\multicolumn{5}{c|}{Decisiveness (\%)} \\ \cline{2-6}\cline{7-11}
          &  \multicolumn{5}{c|}{Words of test data} &  \multicolumn{5}{c|}{Words of test data} \\\cline{2-6}\cline{7-11}
          & 10 & 50 & 100 & 200 & {\bf All} & 10 & 50 & 100 & 200 & {\bf All} \\ \hline
0 & 47.7 & 76.0 & 83.7 & 80.8 & {\bf 72.1} & 36.3 & 38.7 & 39.5 & 42.9 & {\bf 39.3}\\
10 & 50.9 & 86.9 & 96.8 & 99.5 & {\bf 83.5} & 2.13 & 15.5 & 16.5 & 18.1 & {\bf 13.1}\\
12 & 50.9 & 86.9 & 96.8 & 99.7 & {\bf 83.6} & 1.07 & 14.1 & 14.9 & 16.0 & {\bf 11.5}\\
\hline
\end{tabular}
\caption{Performance on genre identification}
\label{tab3}
\end{center}
\end{table*}
Beyond the activation threshold of 12, there is no significant improvement in
accuracy. The technique seems to give good accuracy when there is sufficient
input (100 words or more), but at the cost of very low decisiveness. Excluding
a fixed list of common words such as function words might increase the
decisiveness. These results should be taken with a pinch of salt, as the
notion of genre is not very well-defined, and it is not clear that sections of
the Brown corpus really represent coherent categories, but they may provide a
starting point for further investigation.

\subsection{On decisiveness}

Decisiveness represents the degree to which a unique decision has been made
with a high degree of confidence. In cases where no unique decision has been
made, the range of possibilities will often have been reduced: a category is
only still possible at any stage if its high accumulator value is greater than
the low accumulator value of the best rated category. To illustrate this, the
number of categories which are still possible when all the input was exhausted
was examined. The results appear in tables~\ref{tab4} and ~\ref{tab5}, for the
tests of language identification from word shape tokens with an activation
threshold of 14 and a training set of 2000 tokens, and for genre
identification with a threshold of 12 and a training set of 20000
tokens. Results are shown for the cases of a correct decision, an incorrect
one, and all cases.
\begin{table}[htb]
\begin{center}
\begin{tabular}{|c|r|r|r|}\hline
Languages & \multicolumn{3}{c|}{Number of tests}\\ \cline{2-4}
remaining  & Correct & Incorrect & All \\ \hline
1 & 1560 & 6 & 1566 \\
2 & 128 & 7 & 135 \\
3 & 37 & 9 & 46 \\
4 & 18 & 2 & 20 \\
5 & 5 & 1 & 6 \\
6 & 5 & 1 & 6 \\
7 & 2 & 0 & 2 \\
8 & 2 & 0 & 2 \\
9 & 1 & 0 & 1 \\
10 & 3 & 0 & 3 \\
11 & 1 & 0 & 1 \\
12 & 2 & 0 & 2 \\
13 & 2 & 0 & 2 \\
17 & 1 & 0 & 1 \\
18 & 7 & 0 & 7 \\
\hline
\end{tabular}
\caption{Categories remaining at end of input (language identification from
word shape tokens)}
\label{tab4}
\end{center}
\end{table}
\begin{table}[htb]
\begin{center}
\begin{tabular}{|c|r|r|r|}\hline
Genres & \multicolumn{3}{c|}{Number of tests}\\ \cline{2-4}
remaining  & Correct & Incorrect & All \\ \hline
1 & 173 & 0 & 173 \\
2 & 22 & 0 & 22 \\
3 & 31 & 1 & 32 \\
4 & 45 & 3 & 48 \\
5 & 34 & 4 & 38 \\
6 & 65 & 8 & 73 \\
7 & 73 & 4 & 77 \\
8 & 83 & 3 & 86 \\
9 & 84 & 3 & 87 \\
10 & 89 & 2 & 91 \\
11 & 84 & 0 & 84 \\
12 & 128 & 1 & 129 \\
13 & 131 & 0 & 131 \\
14 & 175 & 1 & 176 \\
15 & 253 & 0 & 253 \\
\hline
\end{tabular}
\caption{Categories remaining at end of input (genre identification)}
\label{tab5}
\end{center}
\end{table}
The average number of possibilities remaining is 1.3 out of 18 for the
language identification test, and 9.7 out of 15 for the genre test, showing
that we are generally near to convergence in the former case, but have only
achieved a small reduction in the possibilities in the latter, in keeping with
the generally low decisiveness.

\subsection{A further comparison}

The classification algorithm described above was originally developed in
response to Sibun and Spitz's work. There is another approach to language
identification, which has a certain amount in common with ours, described in a
patent by Martino and Paulsen~(1996). Their approach is to build tables of the
most frequent words in each language, and assign them a normalised score,
based on the frequency of occurrence of the word in one language compared to
the total across all the languages. Only the most frequent words for each
language are used. The algorithm works by accumulating scores, until a preset
number of words has been read or a minimum score has been reached. They also
apply the technique to genre identification. Since there is a clear
similarity, it is perhaps worth highlighting the differences. In terms of the
algorithm, the most important difference is that no confidence measures are
included. The complexities of splitting the data into different frequency
bands for calculating probabilities are thus avoided, but no
test analogous to overlapping confidence intervals can be applied. Martino and
Paulsen say they obtain a high degree of confidence in the decision after
about 100 words, without saying what the actual success rate is; we can
compare this with around 10 words (or tokens) for convergence here.

\section{Conclusions}

We have examined a simple technique for classifying a stream of input tokens
in which confidence measures are used to determine when a correct decision can
be made. The results in table~\ref{tab1} show that there is a tradeoff between
accuracy and the degree to which the algorithm selects a single language. Not
surprisingly, the amount of training data also affects the performance, with
2000 tokens being adequate for accuracy close to 100\%, and convergence
typically being reached in the first 10 tokens. On a more unconstrained
problem, such as genre identification from words alone, the algorithm performs
less well in both accuracy and decisiveness even with significantly more
training data, and is probably not adequate except as a preprocessor to some
more knowledge intensive technique.

In a sense, language identification is not a very interesting problem. As we
have noted, there are plenty of techniques which work well, each with its own
characteristics and suitability for different application areas. What is
perhaps more important is the way the statistical information has been used
here. When we take a statistical or data-led approach to NLP, there are two
things which can help us trust that the technique is accurate. The first is a
belief that the statistical technique is an adequate model of the underlying
process which ``generates'' the data, using theoretical considerations or some
external source of knowledge to inform this belief. The second is quantitative
evaluation on test data which has been characterised by an outside source (for
example, in the case of part of speech tagging, a corpus which has been
manually annotated, or at least automatically tagged and manually
corrected). The problem with quantitative evaluation is that we do not know
whether it will generalise, so that if we train on one data set, we have only
the theoretical model to reassure that the same model will work on a different 
data set. The idea I have been presenting here is to get the statistical
process itself to provide feedback about itself, through the use of confidence
limits which are themselves based in the statistical model. In doing so, we
hope to avoid presenting a result for which we lack adequate evidence.
\newline
\newline
{\bf Acknowledgement}
\newline
Thanks to Robert Keiller of Canon Research Centre Europe for his advice on
computing accurate statistics.

\section*{Appendix}

Confusion matrix for the case of 2000 lines of training data, token,
threshold=14. An entry in this matrix means that the language on the
horizontal axis was classified as being in the language on the vertical axis
in the indicated number of test samples.

(alb = Albanian, cro = Croatian, dan = Danish, dut = Dutch, eng = English, est 
= Estonian, fre = French, ger = German, ita = Italian, lat = Latin, lit =
Lithuanian, mal = Malay, nor = Norwegian, por = Portugese, ser = Serbian, slo =
Slovenian, spa = Spanish, tur = Turkish. Some of the languages are in a
Romanised form.)

\begin{center}
\begin{tabular}{|c|c|c|c|c|c|c|c|c|c|c|c|c|c|c|c|c|c|c|}\hline
& a & c & d & d & e & e & f & g & i & l & l & m & n & p & s & s & s & t \\
& l & r & a & u & n & s & r & e & t & a & i & a & o & o & e & l & p & u \\
& b & o & n & t & g & t & e & r & a & t & t & l & r & r & r & o & a & r \\
\hline
alb & 100 & & & & & & & & & & & & & & & & & \\
cro & & 96 & & & & & & 1 & & & & & 1 & & & & & \\
dan & & & 100 & & & & & & 1 & & & & 1 & & & & & \\
dut & & & & 100 & & & & & & 1 & & & & & & & 1 & \\
eng & & & & & 99 & 2 & & & & & & & & & & & 1 & \\
est & & & & & & 93 & & 2 & & & 1 & & & & & 1 & & \\
fre & & & & & & & 99 & & & & & & & & & & 1 & \\
ger & & & & & 1 & & & 97 & 1 & & & & & & & & & \\
ita & & & & & & 2 & & & 97 & & & & & & & & & \\
lat & & & & & & 1 & & & & 99 & & & & & & & & \\
lit & & & & & & 1 & & & & & 98 & & & & & & & \\
mal & & & & & & & & & & & & 100 & & & & & & \\
nor & & & & & & & & & & & 1 & & 98 & 1 & & & & \\
por & & & & & & & & & & & & & & 98 & & & & \\
ser & & 3 & & & & & & & & & & & & & 100 & 1 & & 1\\
slo & & 1 & & & & & & & 1 & & & & & & & 98 & & \\
spa & & & & & & 1 & 1 & & & & & & & 1 & & & 97 & \\
tur & & & & & & & & & & & & & & & & & & 99\\
\hline
\end{tabular}
\end{center}

\end{document}